\documentclass[11pt,a4paper]{article}
\usepackage[hyperref]{acl2021}
\usepackage{times}
\usepackage{latexsym}
\usepackage{mathrsfs}
\usepackage{graphicx}
\usepackage{subfigure}
\usepackage{CJK}
\usepackage{amsmath}
\usepackage{booktabs}
\usepackage{multirow}

\usepackage{xurl}

\usepackage{microtype}

\aclfinalcopy 
\def\aclpaperid{1840} 

\title{Prevent the Language Model from being Overconfident in \\ Neural Machine Translation}

\author{
  Mengqi Miao\textsuperscript{1}\thanks{\ \ Equal contribution. This work was done when Mengqi Miao was interning at Pattern Recognition Center, WeChat AI, Tencent Inc, China.} \ ,
  Fandong Meng\textsuperscript{2}$^{*}$, 
  Yijin Liu\textsuperscript{2}, 
  Xiao-Hua Zhou\textsuperscript{3}\thanks{ \ \ Corresponding author.}  \ ,
  and Jie Zhou\textsuperscript{2} \\
  \textsuperscript{1}Peking University, China \\
  \textsuperscript{2}Pattern Recognition Center, WeChat AI, Tencent Inc, China \\
  \textsuperscript{3}Beijing International Center for Mathematical Research, \\National Engineering Lab for Big Data Analysis and Applications, \\Department of Biostatistics, Peking University, Beijing, China \\
  \texttt{miaomq@pku.edu.cn},  \texttt{\{fandongmeng, yijinliu\}@tencent.com}\\
   \texttt{azhou@math.pku.edu.cn}, \texttt{withtomzhou@tencent.com} \\
  \\
}

\date{}

\begin{document}
\maketitle
\begin{abstract}
The Neural Machine Translation (NMT) model is essentially a joint language model conditioned on both the source sentence and partial translation. Therefore, the NMT model naturally involves the mechanism of the Language Model (LM) that predicts the next token only based on partial translation. Despite its success, NMT still suffers from the hallucination problem, generating fluent but inadequate translations. The main reason is that NMT pays excessive attention to the partial translation while neglecting the source sentence to some extent, namely overconfidence of the LM. Accordingly, we define the \emph{Margin between the NMT and the LM}, calculated by subtracting the predicted probability of the LM from that of the NMT model for each token. The \emph{Margin} is negatively correlated to the overconfidence degree of the LM. Based on the property, we propose a \emph{Margin}-based Token-level Objective (MTO) and a \emph{Margin}-based Sentence-level Objective (MSO) to maximize the \emph{Margin} for preventing the LM from being overconfident. Experiments on WMT14 English-to-German, WMT19 Chinese-to-English, and WMT14 English-to-French translation tasks demonstrate the effectiveness of our approach, with 1.36, 1.50, and 0.63 BLEU improvements, respectively, compared to the Transformer baseline. The human evaluation further verifies that our approaches improve translation adequacy as well as fluency. \footnote{Code is available at \url{https://github.com/Mlair77/nmt_adequacy}}

\end{abstract}

\section{Introduction}

Neural Machine Translation (NMT) has achieved great success in recent years \cite{seq2seq-2014, cho-etal-2014-learning, bahdanau2014neural, luong-etal-2015-effective, vaswani2017attention, meng2019dtmt, zhangetal2019bridging, yanetal2020multi}, which generates accurate and fluent translation through modeling the next word conditioned on both the source sentence and partial translation. However, NMT faces the hallucination problem, i.e., translations are fluent but inadequate to the source sentences. One important reason is that the NMT model pays excessive attention to the partial translation to ensure fluency while failing to translate some segments of the source sentence~\cite{weng-etal-2020-towards}, which is actually the overconfidence of the Language Model (LM). In the rest of this paper, the LM mentioned refers to the LM mechanism involved in NMT.

Many recent studies attempt to deal with the inadequacy problem of NMT from two main aspects. One is to improve the architecture of NMT, such as adding a coverage vector to track the attention history~\cite{tu2016modeling}, enhancing the cross-attention module~\cite{meng2016interactive,meng2018neural,weng-etal-2020-towards}, and dividing the source sentence into past and future parts~\cite{zheng-etal-2019-dynamic}. The other aims to propose a heuristic adequacy metric or objective based on the output of NMT. \citet{Tu:reconstruct:2017:AAAI} and \citet{Kong_Tu_Shi_Hovy_Zhang_2019} enhance the model's reconstruction ability and increase the coverage ratio of the source sentences by translations, respectively. Although some researches~\cite{Tu:reconstruct:2017:AAAI, Kong_Tu_Shi_Hovy_Zhang_2019, weng-etal-2020-towards} point out that the lack of adequacy is due to the overconfidence of the LM, unfortunately, they do not propose effective solutions to the overconfidence problem.

From the perspective of preventing the overconfidence of the LM, we first define an indicator of the overconfidence degree of the LM, called the \emph{Margin between the NMT and the LM}, by subtracting the predicted probability of the LM from that of the NMT model for each token. A small \emph{Margin} implies that the NMT might concentrate on the partial translation and degrade into the LM, i.e., the LM is overconfident. Accordingly, we propose a \emph{\textbf{M}argin}-based \textbf{T}oken-level \textbf{O}bjective (MTO) to maximize the \emph{Margin}. Furthermore, we observe a phenomenon that if target sentences in the training data contain many words with negative \emph{Margin}, they always do not correspond to the source sentences. These data are harmful to model performance. Therefore, based on the MTO, we further propose a \emph{\textbf{M}argin}-based \textbf{S}entence-level \textbf{O}bjective (MSO) by adding a dynamic weight function to alleviate the negative effect of these ``dirty data''.

We validate the effectiveness and superiority of our approaches on the Transformer~\cite{vaswani2017attention}, and conduct experiments on large-scale WMT14 English-to-German, WMT19 Chinese-to-English, and WMT14 English-to-French translation tasks.
Our contributions are:
\begin{itemize}
\setlength{\topsep}{1pt}
\setlength{\itemsep}{1pt}
\setlength{\parsep}{1pt}
\setlength{\parskip}{1pt}
    \item We explore the connection between inadequacy translation and the overconfidence of the LM in NMT, and thus propose an indicator of the overconfidence degree, i.e., the \emph{Margin} between the NMT and the LM.
    \item Furthermore, to prevent the LM from being overconfident, we propose two effective optimization objectives to maximize the \emph{Margin}, i.e., the Margin-based Token-level Objective (MTO) and the Margin-based Sentence-level Objective (MSO).
    \item Experiments on WMT14 English-to-German, WMT19 Chinese-to-English, and WMT14 English-to-French show that our approaches bring in significant improvements by +1.36, +1.50, +0.63 BLEU points, respectively. Additionally, the human evaluation verifies that our approaches can improve both translation adequacy and fluency. 
\end{itemize}

\section{Background}
Given a source sentence $\mathbf{x}=\{x_1, x_2, ..., x_{_N}\}$, the NMT model predicts the probability of a target sentence $\mathbf{y}=\{y_1, y_2, ..., y_{_T}\}$ word by word:
\begin{equation}
\setlength{\abovedisplayskip}{3pt}
\label{eq:nmt-prob}
    P(\mathbf{y}|\mathbf{x}) = \prod_{t=1}^{T} p(y_t|\mathbf{y}_{<t},\mathbf{x}),
\setlength{\belowdisplayskip}{3pt}
\end{equation}
where $\mathbf{y}_{<t}=\{y_1, y_2, ..., y_{t-1}\}$ is the partial translation before $y_t$. From Eq.~\ref{eq:nmt-prob}, the source sentence $\mathbf{x}$ and partial translation $\mathbf{y}_{<t}$ are considered in the meantime, suggesting that the NMT model is essentially a joint language model and the LM is instinctively involved in NMT.

Based on the encoder-decoder architecture, the encoder of NMT maps the input sentence $\mathbf{x}$ to hidden states. At time step $t$, the decoder of NMT employs the output of the encoder and $\mathbf{y}_{<t}$ to predict $y_t$. The training objective of NMT is to minimize the negative log-likelihood, which is also known as the cross entropy loss function:
\begin{equation}
\setlength{\abovedisplayskip}{3pt}
\label{eq:cross-entropy}
    \mathcal{L}_{ce}^{NMT}= - \sum_{t=1}^{T} log \ p(y_t|\mathbf{y}_{<t},\mathbf{x}).
\setlength{\belowdisplayskip}{3pt}
\end{equation}

The LM measures the probability of a target sentence similar to NMT but without knowledge of the source sentence $\mathbf{x}$:
\begin{equation}
\setlength{\abovedisplayskip}{3pt}
    P(\mathbf{y}) = \prod_{t=1}^{T} p(y_t|\mathbf{y}_{<t}).
\setlength{\belowdisplayskip}{3pt}
\end{equation}
The LM can be regarded as the part of NMT decoder that is responsible for fluency, only takes $y_{<t}$ as input.
The training objective of the LM is almost the same as NMT except for the source sentence $\mathbf{x}$:
\begin{equation}
\setlength{\abovedisplayskip}{3pt}
\label{eq:lm_ce_loss}
    \mathcal{L}_{ce}^{LM}= - \sum_{t=1}^{T} log \ p(y_t|\mathbf{y}_{<t}).
\setlength{\belowdisplayskip}{3pt}
\end{equation}

The NMT model predicts the next word $y_t$ according to the source sentence $\mathbf{x}$ and meanwhile ensures that $y_t$ is fluent with the partial translation $\mathbf{y}_{<t}$. However, when NMT pays excessive attention to translation fluency, some source segments may be neglected, leading to inadequacy problem. This is exactly what we aim to address in this paper.

\section{The Approach} \label{sec:method}
In this section, we firstly define the \emph{Margin between the NMT and the LM} (Section \ref{sec:margin_define}), which reflects the overconfidence degree of the LM. Then we put forward the token-level (Section \ref{sec:token_level}) and sentence-level (Section \ref{sec:sentence_level}) optimization objectives to maximize the \emph{Margin}. Finally, we elaborate our two-stage training strategy  (Section \ref{sec:training}).

\subsection{\emph{Margin} between the NMT and the LM}
\label{sec:margin_define}
When the NMT model excessively focuses on partial translation, i.e., the LM is overconfident, the NMT model degrades into the LM, resulting in hallucinated translations. To prevent the overconfidence problem, we expect that the NMT model outperforms the LM as much as possible in predicting golden tokens. Consequently, we define the \emph{Margin between the NMT and the LM} at the $t$-th time step by the difference of the predicted probabilities of them:
\begin{equation}
    \label{eq:delta_def}
    \Delta(t) = p_{_{NMT}}(y_t|\mathbf{y}_{<t},\mathbf{x})  - p_{_{LM}}(y_t|\mathbf{y}_{<t}),
\end{equation}
where $p_{_{NMT}}$ denotes the predicted probability of the NMT model, i.e.,  $p(y_t|\mathbf{y}_{<t},\mathbf{x})$, and $p_{_{LM}}$ denotes that of the LM, i.e., $p(y_t|\mathbf{y}_{<t})$.

The \emph{Margin} $\Delta(t)$ is negatively correlated to the overconfidence degree of the LM, and different values of the \emph{Margin} indicate different cases:

\begin{itemize}
\label{item:delta-means}
\setlength{\topsep}{1pt}
\setlength{\itemsep}{1pt}
\setlength{\parsep}{1pt}
\setlength{\parskip}{1pt}
    \item If $\Delta(t)$ is big, the NMT model is apparently better than the LM, and $y_t$ is strongly related to the source sentence $\mathbf{x}$. Hence the LM is not overconfident. 
    \item If $\Delta(t)$ is medium, the LM may be slightly overconfident and the NMT model has the potential to be enhanced. 
    \item If $\Delta(t)$ is small, the NMT model might degrade to the LM and not correctly translate the source sentence, i.e., the LM is overconfident.\footnote{In addition, if $p_{_{NMT}}(y_t|\mathbf{y}_{<t},\mathbf{x})$ is large, less attention will be paid to this data because $y_t$ has been learned well, which will be described in detail in Section~\ref{sec:token_level}.} 
\end{itemize}
Note that sometimes, the model needs to focus more on the partial translation such as the word to be predicted is a determiner in the target language. In this case, although small $\Delta(t)$ does not indicate the LM is overconfident, enlarging the $\Delta(t)$ can still enhance the NMT model.

\subsection{\emph{Margin}-based Token-level Objective}
\label{sec:token_level}
Based on the \emph{Margin}, we firstly define the \emph{Margin} loss $\mathcal{L}_{M}$ and then fuse it into the cross entropy loss function to obtain the \emph{\textbf{M}argin}-based \textbf{T}oken-evel Optimization \textbf{O}bjective (MTO).
Formally, we define the \emph{Margin} loss $\mathcal{L}_{M}$ to maximize the \emph{Margin} as follow:
\begin{equation}
    \label{eq:margin_term}
    \mathcal{L}_{M} = \sum_{t=1}^{T} (1 - p_{_{NMT}}(t)) \mathcal{M}(\Delta(t)),
\end{equation}
where we abbreviate $p_{_{NMT}}(y_t|\mathbf{y}_{<t},\mathbf{x})$ as $p_{_{NMT}}(t)$. $\mathcal{M}(\Delta(t))$ is a function of $\Delta(t)$, namely \emph{Margin} function, which is monotonically decreasing (e.g., $1-\Delta(t)$). Moreover, when some words have the same $\Delta(t)$ but different $p_{_{NMT}}(t)$, their meanings are quite different: (1) If $p_{_{NMT}}(t)$ is big, the NMT model learns the token well and does not need to focus on the \emph{Margin} too much; (2) If $p_{_{NMT}}(t)$ is small, the NMT model is urgently to be optimized on the token thus the weight of $\mathcal{M}(\Delta(t))$ should be enlarged. Therefore, as the weight of $\mathcal{M}(\Delta(t))$, $1 - p_{_{NMT}}(t)$ enables the model treat tokens wisely.

\paragraph{Variations of $\mathcal{M}(\Delta)$.}
\label{four-M-delta-function}
We abbreviate \emph{Margin} function $\mathcal{M}(\Delta(t))$ as $\mathcal{M}(\Delta)$ hereafter. A simple and intuitive definition is the \emph{Linear} function: $\mathcal{M}(\Delta) = 1 - \Delta$, which has the same gradient for different $\Delta$. However, as illustrated in Section \ref{sec:margin_define}, different $\Delta$ has completely various meaning and needs to be treated differently. Therefore, we propose three non-linear \emph{Margin} functions $\mathcal{M}(\Delta)$ as follows:
\begin{itemize}
\setlength{\itemsep}{1pt}
\setlength{\parsep}{1pt}
\setlength{\parskip}{1pt}
    \item \emph{Cube}: $(1- \Delta^3) /2 $.
    \item \emph{Quintic} (fifth power): $(1- \Delta^5) /2 $.
    \item \emph{Log}: $\frac{1}{\alpha}log(\frac{1-\Delta}{1+\Delta}) + 0.5$.
\end{itemize}
where $\alpha$ is a hyperparamater for \emph{Log}.

\begin{figure}[t!]
    \centering
    \includegraphics[width=0.45\textwidth]{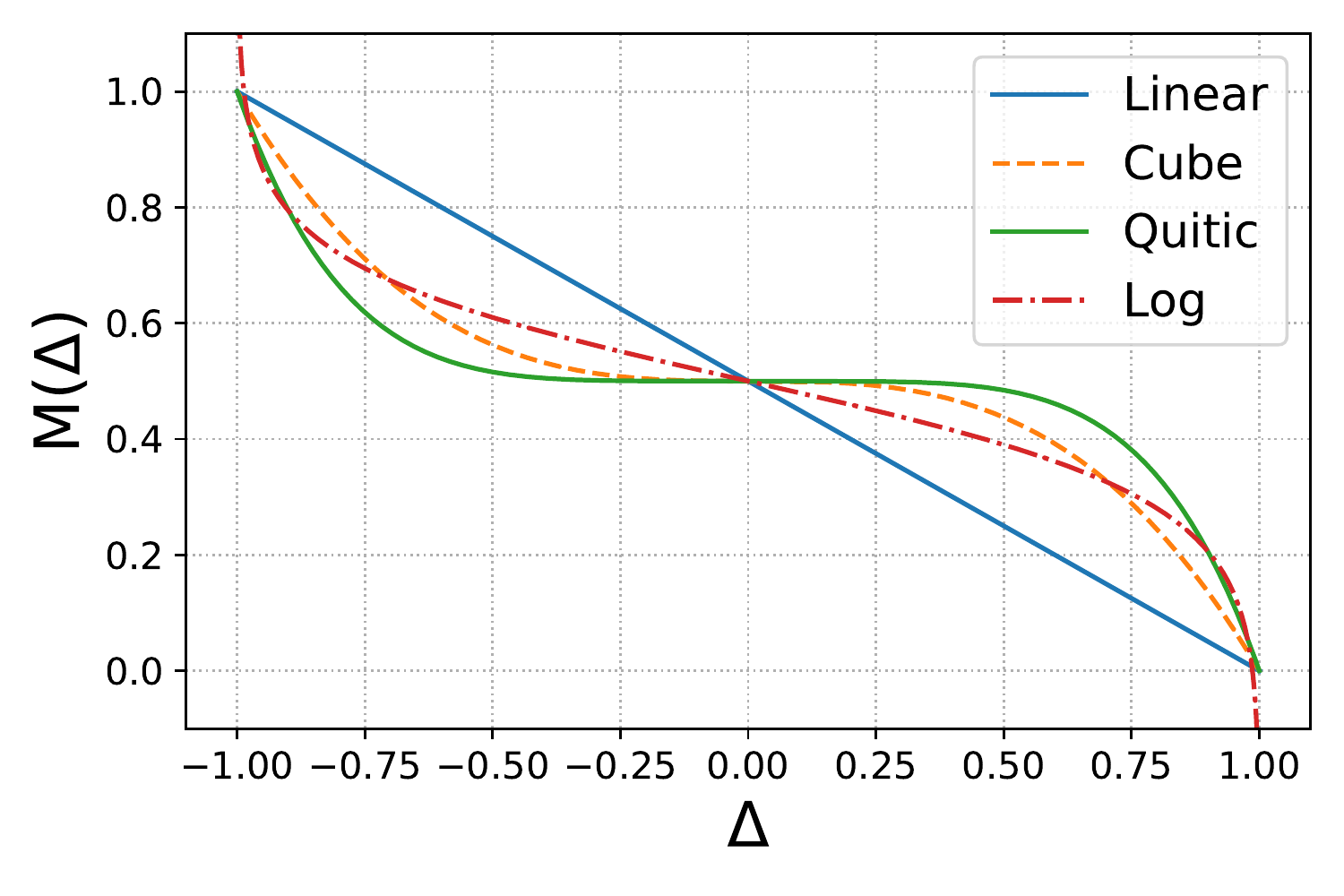}
    \caption{The four \emph{Margin} functions $\mathcal{M}(\Delta)$. All of them are monotonically decreasing, yet with different slopes. Compared with \emph{Linear}, the three non-linear functions are more stable around $|\Delta|=0$ and steeper around $|\Delta|=1$. We set $\alpha$ in \emph{Log} to 10 in this figure.}
    \label{fig:nonlinear}
\end{figure}

As shown in Figure \ref{fig:nonlinear}, the four variations\footnote{In order to keep the range of $\mathcal{M}(\Delta)$ roughly [0,1], we set \emph{Linear} function to $(1 - \Delta)/2$.} have quite different slopes. Specifically, the three non-linear functions are more stable around $\Delta=0$ (e.g., $\Delta \in [-0.5, 0.5]$) than \emph{Linear}, especially \emph{Quintic}. We will report the performance of the four $\mathcal{M}(\Delta)$ concretely and analyze why the three non-linear $\mathcal{M}(\Delta)$ perform better than \emph{Linear} in Section \ref{sec:analysis}. 

Finally, based on $\mathcal{L}_{M}$, we propose the \emph{\textbf{M}argin}-based \textbf{T}oken-level \textbf{O}bjective (MTO):
\begin{equation}
\label{eq:token_loss}
    \mathcal{L}^{T}= \mathcal{L}_{ce}^{NMT} + \lambda_{M} \mathcal{L}_{M},
\end{equation}
where $\mathcal{L}_{ce}^{NMT}$ is the cross-entropy loss of the NMT model defined in Eq. \ref{eq:cross-entropy} and $\lambda_{M}$ is the hyperparameter for the \emph{Margin} loss $\mathcal{L}_{M}$.

\subsection{\emph{Margin}-based Sentence-level Objective}
\label{sec:sentence_level}
Furthermore, through analyzing the \emph{Margin} distribution of target sentences, we observe that the target sentences in the training data which have many tokens with negative \emph{Margin} are almost ``hallucinations'' of the source sentences (i.e., dirty data), thus will harm the model performance. Therefore, based on MTO, we further propose the \emph{\textbf{M}argin}-based \textbf{S}entence-level \textbf{O}bjective (MSO) to address this issue.

\begin{figure}[tbp!]
    \centering
    \includegraphics[width=0.47\textwidth]{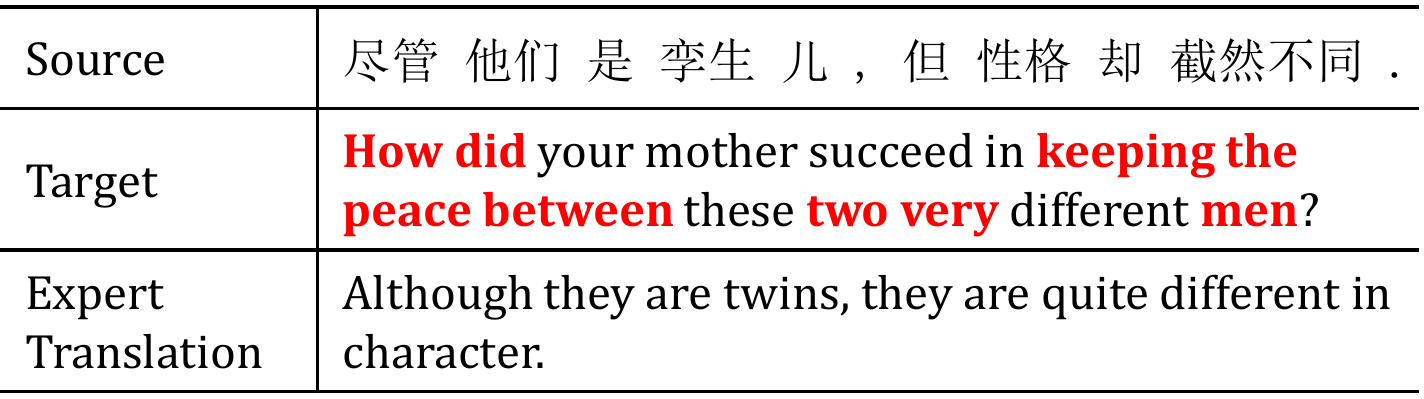}
    \caption{The parallel sentences, i.e., the source and target sentences, are sampled from the WMT19 Chinese-to-English training dataset. We also list an expert translation of the source sentence. The words in bold red have negative \emph{Margin}. This target sentence has more than 50\% tokens with negative \emph{Margin}, and these tokens are almost irrelevant to the source sentence. Apparently, the target sentence is a hallucination and will harm the model performance.}
    \label{fig:case_delta0}
\end{figure}

Compared with the LM, the NMT model predicts the next word with more prior knowledge (i.e., the source sentence). Therefore, it is intuitive that when predicting $y_t$, the NMT model should predict more accurately than the LM, as follow:
\begin{equation}
\label{eq:delta_0}
\setlength{\abovedisplayskip}{7pt}
\setlength{\belowdisplayskip}{7pt}
p_{_{NMT}}(y_t|\mathbf{y}_{<t},\mathbf{x}) > p_{_{LM}}(y_t|\mathbf{y}_{<t}).
\end{equation}
Actually, the above equation is equivalent to $\Delta(t) > 0$. The larger $\Delta(t)$ is, the more the NMT model exceeds the LM. However, there are many tokens with negative \emph{Margin} through analyzing the \emph{Margin} distribution. We conjecture the reason is that the target sentence is not corresponding to the source sentence in the training corpus, i.e., the target sentence is a hallucination.
Actually, we also observe that if a large proportion of tokens in a target sentence have negative \emph{Margin} (e.g., 50\%), the sentence is probably not corresponding to the source sentence, such as the case in Figure \ref{fig:case_delta0}. These ``dirty'' data will harm the performance of the NMT model.

To measure the ``dirty'' degree of data, we define the Sentence-level Negative Margin Ratio of parallel sentences $(\mathbf{x}, \mathbf{y})$ as follow: 
\begin{equation}
\setlength{\abovedisplayskip}{7pt}
\setlength{\belowdisplayskip}{7pt}
    \label{eq:sen-score}
    \mathcal{R}(\mathbf{x}, \mathbf{y}) = \frac{\#\{y_t \in \mathbf{y}: \Delta(t) < 0 \}}{\#\{y_t: y_t \in \mathbf{y}\}},
\end{equation}
where $\#\{y_t\in\mathbf{y}: \Delta(t) < 0 \}$ denotes the number of tokens with negative $\Delta(t)$ in $\mathbf{y}$, and $\#\{y_t: y_t\in\mathbf{y}\}$ is the length of the target sentence $\mathbf{y}$.

When $ \mathcal{R} (\mathbf{x},\mathbf{y})$ is larger than a threshold $k$ (e.g., $k$=50\%), the target sentence may be desperately inadequate, or even completely unrelated to the source sentence, as shown in Figure~\ref{fig:case_delta0}. In order to eliminate the impact of these seriously inadequate sentences, we ignore their loss during training by the \emph{\textbf{M}argin}-based \textbf{S}entence-level \textbf{O}bjective (MSO):
\begin{equation}
\setlength{\abovedisplayskip}{7pt}
\setlength{\belowdisplayskip}{7pt}
\label{eq:sentence-loss}
    \mathcal{L}^{S}=I_{\mathcal{R} (\mathbf{x}, \mathbf{y})< k} \cdot \mathcal{L}^{T},
\end{equation}
where $I_{\mathcal{R} (\mathbf{x}, \mathbf{y})< k}$ is a dynamic weight function in sentence level. The indicative function $I_{\mathcal{R} (\mathbf{x}, \mathbf{y})< k}$ equals to 1 if $\mathcal{R} (\mathbf{x}, \mathbf{y})< k $, else 0, where $k$ is a hyperparameter. $\mathcal{L}^{T}$ is MTO defined in Eq.~\ref{eq:token_loss}.

$I_{\mathcal{R} (\mathbf{x}, \mathbf{y})< k}$ is dynamic at the training stage. During training, as the model gets better, its ability to distinguish hallucinations improves thus $I_{\mathcal{R} (\mathbf{x}, \mathbf{y})< k}$ becomes more accurate. We will analyze the changes of $I_{\mathcal{R}(\mathbf{x}, \mathbf{y})< k}$ in Section \ref{sec:analysis}.

\subsection{Two-stage Training}
\label{sec:training}
We elaborate our two-stage training in this section, 1) jointly pretraining an NMT model and an auxiliary LM, and 2) finetuning the NMT model.

\paragraph{Jointly Pretraining.}
The language model mechanism in NMT cannot be directly evaluated, thus we train an auxiliary LM to represent it. We pretrain them together using a fusion loss function:
\begin{equation}
\setlength{\abovedisplayskip}{7pt}
\setlength{\belowdisplayskip}{7pt}
    \label{eq:joint-training}
    \mathcal{L}_{pre} = \mathcal{L}^{NMT}_{ce} + \lambda_{LM} \mathcal{L}^{LM}_{ce},
\end{equation}
where $\mathcal{L}^{NMT}_{ce}$ and $\mathcal{L}^{LM}_{ce}$ are the cross entropy loss functions of the NMT model and the LM defined in Eq.~\ref{eq:cross-entropy} and Eq.~\ref{eq:lm_ce_loss}, respectively. $\lambda_{LM}$ is a hyperparameter. Specifically, we jointly train them through sharing their decoders' embedding layers and their pre-softmax linear transformation layers~\cite{vaswani2017attention}. There are two reasons for joint training: (1) making the auxiliary LM as consistent as possible with the language model mechanism in NMT; (2) avoiding abundant extra parameters.

\paragraph{Finetuning.}
We finetune the NMT model by minimizing the MTO ($\mathcal{L}^{T}$ in Eq.~\ref{eq:token_loss}) and MSO ($\mathcal{L}^{S}$ in Eq.~\ref{eq:sentence-loss}).\footnote{The LM can be fixed or trained along with the NMT after pretraining. Our experimental results show that continuous training the LM and fixing the LM have analogous performance during the finetuning stage. Therefore, we only report the results of keeping the LM fixed in this paper.}
Note that the LM is not involved at the inference stage.

\section{Experimental Settings}

We conduct experiments on three large-scale NMT tasks, i.e., WMT14 English-to-German (En$\to$De), WMT14 English-to-French (En$\to$Fr), and WMT19 Chinese-to-English (Zh$\to$En).

\paragraph{Datasets.} 
For En$\to$De, we use 4.5M training data. Following the same setting in \cite{vaswani2017attention}, we use \emph{newstest2013} as validation set and \emph{newstest2014} as test set, which contain 3000 and 3003 sentences, respectively. For En$\to$Fr, the training dataset contains about 36M sentence pairs, and we use \emph{newstest2013} with 3000 sentences as validation set and \emph{newstest2014} with 3003 sentences as test set. For Zh$\to$En, we use 20.5M training data and use \emph{newstest2018} as validation set and \emph{newstest2019} as test set, which contain 3981 and 2000 sentences, respectively. For Zh$\to$En, the number of merge operations in byte pair encoding (BPE) ~\cite{bpe-sennrich-etal-2016-neural} is set to 32K for both source and target languages. For En$\to$De and En$\to$Fr, we use a shared vocabulary generated by 32K BPEs. 

\paragraph{Evaluation.}
We measure the case-sensitive BLEU scores using \emph{multi-bleu.perl} \footnote{\url{https://github.com/moses-smt/mosesdecoder/blob/master/scripts/generic/multi-bleu.perl}} for En$\to$De and En$\to$Fr. For Zh$\to$En, case-sensitive BLEU scores are calculated by Moses \emph{mteval-v13a.pl} script\footnote{\url{https://github.com/moses-smt/mosesdecoder/blob/mast-er/scripts/generic/mteval-v13a.pl}}. Moreover, we use the paired bootstrap resampling~\cite{koehn-2004-statistical} for significance test. We select the model which performs the best on the validation sets and report its performance on the test sets for evaluation.

\paragraph{Model and Hyperparameters.}
We conduct experiments based on the Transformer~\cite{vaswani2017attention} and implement our approaches with the open-source tooklit \emph{Opennmt-py}~\cite{klein-etal-2017-opennmt}. Following the Transformer-Base setting in~\cite{vaswani2017attention}, we set the hidden size to 512 and the encoder/decoder layers to 6. All three tasks are trained with 8 NVIDIA V100 GPUs, and the batch size for each GPU is 4096 tokens. The beam size is 5 and the length penalty is 0.6. Adam optimizer~\cite{kingma2014adam} is used in all the models. 
The LM architecture is the decoder of the Transformer excluding the cross-attention layers, sharing the embedding layer and the pre-softmax linear transformation with the NMT model. For En$\to$De, Zh$\to$En, and En$\to$Fr, the number of training steps is 150K for jointly pretraining stage and 150K for finetuning\footnote{The LM does not need to be state-of-the-art. The previous study of \cite{lm_prior:2020:EMNLP} has shown that a more powerful LM does not lead to further improvements to NMT.}. During pretraining, we set $\lambda_{LM}$ to 0.01 for all three tasks\footnote{The experimental results show that the model is insensitive to $\lambda_{LM}$. Therefore we make $\lambda_{LM}$ consistent for all the three tasks.}. Experimental results shown in Appendix~\ref{sec:lm_loss} indicate that the LM has converged after pretraining for all the three tasks. During finetuning, the \emph{Margin} function $\mathcal{M}(\Delta)$ in Section~\ref{sec:token_level} is set to \emph{Quintic}, and we will analyze the four $\mathcal{M}(\Delta)$ in Section~\ref{sec:analysis}. $\lambda_{M}$ in Eq.~\ref{eq:token_loss} is set to 5, 8, and 8 on En$\to$De, En$\to$Fr and Zh$\to$En, respectively. For MSO, the threshold $k$ in Eq.~\ref{eq:sentence-loss} is set to 30\% for En$\to$De and Zh$\to$En, 40\% for En$\to$Fr. The two hyperparameters (i.e., $\lambda_{M}$ and $k$) are searched on validation sets, and the selection details are shown in Appendix~\ref{sec:hyper-select}.
The baseline model (i.e., vanilla Transformer) is trained for 300k steps for En$\to$De, En$\to$Fr and Zh$\to$En. Moreover, we use a joint training model as our secondary baseline, namely NMT+LM, by jointly training the NMT model and the LM throughout the training stage with 300K steps. The training steps of all the models are consistent, thus the experiment results are strictly comparable.

\begin{table}[!t]
    \centering
    \renewcommand{\arraystretch}{1.1}
    \setlength{\tabcolsep}{1mm}{
    \scalebox{0.85}{
    \begin{tabular}{ l | c | c }
     \toprule
    \textbf{System} & \textbf{En$\to$De} &  $\uparrow$ \\ 
    \hline
    \multicolumn{3}{c}{\em Existing NMT systems} \\
    \hline
    Transformer~\cite{vaswani2017attention} & 27.3 & - \\
    MRT*~\cite{MRT-2016} & 27.71 & -\\
    Simple Fusion**~\cite{simple_fusion_2018} & 27.88 & - \\
    Localness~\cite{yang-etal-2018-localness} & 28.11 & - \\
    Context-Aware~\cite{yang2019contextaware} & 28.26 & - \\
    AOL~\cite{Kong_Tu_Shi_Hovy_Zhang_2019} & 28.01 & -   \\
    Eval. Module~\cite{feng2019fluency_faithfulness} & 27.55 & - \\
    Past\&Future~\cite{zheng-etal-2019-dynamic} & 28.10 & - \\
    Dual~\cite{yan2020dual} & 27.86 & - \\
    Multi-Task~\cite{weng-etal-2020-towards} & 28.25  & - \\
    \hline
    \multicolumn{3}{c}{\em Our NMT systems} \\
    \hline 
    NMT (Transformer)  & 27.22 & ref  \\
    ~~~~+ LM            &  27.97 & +0.75  \\ 
    ~~~~~~~+ MTO  &  28.47$^{\dagger\ddagger}$ & +1.25  \\
    ~~~~~~~+ MSO  & \textbf{28.58}$^{\dagger\ddagger}$ & +1.36 \\
    \bottomrule
    \end{tabular}}}    
    \caption{Case-sensitive BLEU scores (\%) on the test set of WMT14 En$\to$De. $\uparrow$ denotes the improvement compared with the NMT baseline (i.e., Transformer). ``$\dagger$'': significantly better than NMT ($p$\textless0.01). ``$\ddagger$'': significantly better than the joint model NMT+LM ($p$\textless0.01). (MRT* in~\cite{MRT-2016} is RNN-based, and the result reported here is implemented on Transformer by \citet{weng-etal-2020-towards}. **: we re-implement Simple Fusion on upon of Transformer.)}
    \label{tab:ende}
\end{table}

\section{Results and Analysis}
We first evaluate the main performance of our approaches (Section \ref{sec:ende_results} and~\ref{sec:enfr_zhen_results}). Then, the human evaluation further confirms the improvements of translation adequacy and fluency (Section~\ref{sec:human_eval}). Finally, we analyze the positive impact of our models on the distribution of \emph{Margin} and explore how each fragment of our method works (Section~\ref{sec:analysis}).

\subsection{Results on En$\to$De}
\label{sec:ende_results}
The results on WMT14 English-to-German (En$\to$De) are summarized in Table \ref{tab:ende}. We list the results from \cite{vaswani2017attention} and several related competitive NMT systems by various methods, such as Minimum Risk Training (MRT) objective~\cite{MRT-2016}, Simple Fusion of NMT and LM~\cite{simple_fusion_2018}, optimizing adequacy metrics~\cite{Kong_Tu_Shi_Hovy_Zhang_2019,feng2019fluency_faithfulness} and improving the Transformer architecture~\cite{yang-etal-2018-localness, zheng-etal-2019-dynamic, yang2019contextaware, weng-etal-2020-towards, yan2020dual}. We re-implement the Transformer model~\cite{vaswani2017attention} as our baseline. Similarly, we re-implement the Simple Fusion~\cite{simple_fusion_2018} model.~\footnote{The architectures of the LM and NMT model in Simple Fusion are consistent with our MTO and MSO.} 
Finally, the results of the joint training model NMT+LM, and models with our MTO and MSO objectives are reported.

Compared with the baseline, NMT+LM yields +0.75 BLEU improvement. Based on NMT+LM, our MTO achieves further improvement with +0.50 BLEU scores, indicating that preventing the LM from being overconfident could significantly enhance model performance. Moreover, MSO performs better than MTO by +0.11 BLEU scores, which implies that the ``dirty data'' in the training dataset indeed harm the model performance, and the dynamic weight function $I_{\mathcal{R} (\mathbf{x}, \mathbf{y})< k}$ in Eq.~\ref{eq:sentence-loss} could reduce the negative impact. In conclusion, our approaches improve up to +1.36 BLEU scores on En$\to$De compared with the Transformer baseline and substantially outperforms the existing NMT systems. The results demonstrate the effectiveness and superiority of our approaches.

\begin{table}[!t]
    \centering
    \renewcommand{\arraystretch}{1.1}
    \setlength{\tabcolsep}{1mm}{
    \scalebox{0.85}{
    \begin{tabular}{ l | c c | c c }
     \toprule
    \multirow{2}{*}{\textbf{System}} & \multicolumn{2}{c|}{\textbf{En$\to$Fr}} & \multicolumn{2}{c}{\textbf{Zh$\to$En}}\\ \cline{2-5} 
                             & BLEU       & $\uparrow$       & BLEU   & $\uparrow$  \\ 
                            \hline
    \citet{vaswani2017attention}* & 38.1 & - &  - & - \\
    \hline
    NMT (Transformer)   & 41.07  & ref & 25.75 & ref \\
    ~~~~+ LM          & 41.14 & +0.07 & 25.90 & +0.15 \\ 
    ~~~~~~~+ MTO      & 41.56$^{\dagger\ddagger}$ & +0.49 & 26.94$^{\dagger\ddagger}$ & +1.19 \\
    ~~~~~~~+ MSO   & \textbf{41.70}$^{\dagger\ddagger}$ & +0.63 & \textbf{27.25}$^{\dagger\ddagger}$ & +1.50 \\
    \bottomrule
    \end{tabular}}}    
    \caption{Case-sensitive BLEU scores (\%) on the test set of WMT14 En$\to$Fr and WMT19 Zh$\to$En. $\uparrow$ denotes the improvement compared with the NMT baseline (i.e., Transformer). ``$\dagger$'': significantly better than NMT ($p$\textless0.01). ``$\ddagger$'': significantly better than the joint model NMT+LM ($p$\textless0.01). * denotes the results come from the cited paper.}
    \label{tab:enfr-zhen} 
\end{table}

\subsection{Results on En$\to$Fr and Zh$\to$En}
\label{sec:enfr_zhen_results}
The results on WMT14 English-to-French (En$\to$Fr) and WMT19 Chinese-to-English (Zh$\to$En) are shown in Table~\ref{tab:enfr-zhen}. We also list the results of~\cite{vaswani2017attention} and our reimplemented Transformer as the baselines.

On En$\to$Fr, our reimplemented result is higher than the result of~\cite{vaswani2017attention}, since we update 300K steps while \citet{vaswani2017attention} only update 100K steps. Many studies obtain similar results to ours (e.g., 41.1 BLEU scores from~\cite{ott-etal-2019-fairseq}). Compared with the baseline, NMT+LM yields +0.07 and +0.15 BLEU improvements on En$\to$Fr and Zh$\to$En, respectively. The improvement of NMT+LM on En$\to$De in Table~\ref{tab:ende} (i.e., +0.75) is greater than these two datasets. We conjecture the reason is that the amount of training data of En$\to$De is much smaller than that of En$\to$Fr and Zh$\to$En, thus NMT+LM is more likely to improve the model performance on En$\to$De.

Compared with NMT+LM, our MTO achieves further improvements with +0.42 and +1.04 BLEU scores on En$\to$Fr and Zh$\to$En, respectively, which demonstrates the performance improvement is mainly due to our \emph{Margin}-based objective rather than joint training. Moreover, based on MTO, our MSO further yields +0.14 and +0.31 BLEU improvements. In summary, our approaches improve up to +0.63 and +1.50 BLEU scores on En$\to$Fr and Zh$\to$En compared with the baselines, respectively, which demonstrates the effectiveness and generalizability of our approaches.

\begin{table}[t]
\centering
\scalebox{0.85}{
\begin{tabular}{l|ccc}
\hline
\toprule
\textbf{Model}    & \textbf{Adequacy} & \textbf{Fluency} & \textbf{Ave.}\\  \hline
NMT (Transformer)    & 4.04    & 4.66  & 4.35 \\
~~~~+ LM            & 4.12     & 4.86 & 4.49   \\
~~~~~~~+ MTO   & 4.26     & 4.87  & 4.57   \\
~~~~~~~+ MSO   & \textbf{4.41} & \textbf{4.91}  & \textbf{4.66}  \\
\bottomrule
\end{tabular}}
\caption{Human evaluation on adequacy and fluency.}
\label{table:human-eval}
\end{table}

\subsection{Human Evaluation}
\label{sec:human_eval}
We conduct the human evaluation for translations in terms of adequacy and fluency. Firstly, we randomly sample 100 sentences from the test set of WMT19 Zh$\to$En. Then we invite three annotators to evaluate the translation adequacy and fluency. Five scales have been set up, i.e., {1, 2, 3, 4, 5}. For adequacy, ``1'' means totally irrelevant to the source sentence, and ``5'' means equal to the source sentence semantically. For fluency, ``1'' represents not fluent and incomprehensible; ``5'' represents very ``native''. Finally, we take the average of the scores from the three annotators as the final score. 

The results of the baseline and our approaches are shown in Table \ref{table:human-eval}. Compared with the NMT baseline, NMT+LM, MTO and MSO improve adequacy with 0.08, 0.22, and 0.37 scores, respectively. Most improvements come from our \emph{Margin}-based methods MTO and MSO, and MSO performs the best. For fluency, NMT+LM achieves 0.2 improvement compared with NMT. Based on NMT+LM, MTO and MSO yield further improvements with 0.01 and 0.05 scores, respectively. Human evaluation indicates that our MTO and MSO approaches remarkably improve translation adequacy and slightly enhance translation fluency.

\begin{figure}[tbp!]
    \centering
    \includegraphics[width=0.4\textwidth]{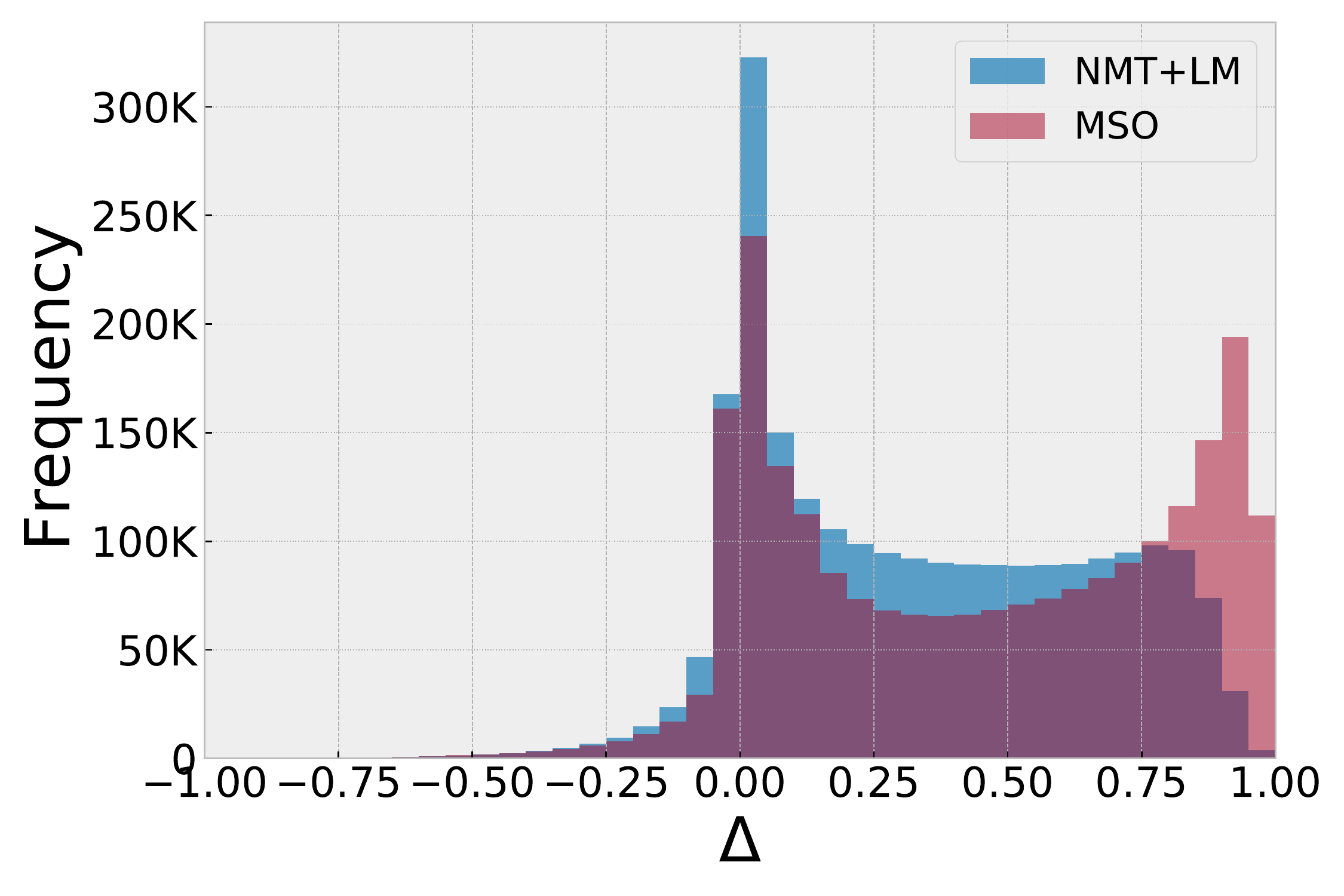}
    \caption{The distribution of $\Delta$ of NMT+LM and MSO. We randomly sample 100K sentence pairs from the training dataset of Zh$\to$En and compute the \emph{Margin} of their tokens. The purple area is the overlap of the two models' $\Delta$ distributions. The two distributions are quite different. Compared with NMT+LM, MSO reduces the distribution around $\Delta=0$ and meanwhile increases the distribution around $\Delta=1$.}
    \label{fig:delta_hist}
\end{figure}

\begin{table}[!tbp]
\centering
\scalebox{0.8}{
\begin{tabular}{l|l|l}
\hline
\toprule
\textbf{Model}    & \textbf{Percent of $\Delta<0~(\downarrow)$} & \textbf{Average $\Delta~(\uparrow)$} \\  \hline
NMT + LM       & 12.45\% (ref)      & 0.33 (ref)\\
~~~~~~~~~~~~~+ MTO    & \textbf{10.17\%} (-2.28\%)    & 0.44 (+0.11)  \\
~~~~~~~~~~~~~+ MSO    & 10.89\% (-1.56\%)  & 0.44 (+0.11)  \\
\bottomrule
\end{tabular}}
\caption{The percent of $\Delta<0$ and average $\Delta$ of models computed from the 100K sentence pairs introduced in Figure~\ref{fig:delta_hist}. Compared with NMT+LM, both MTO and MSO effectively reduce the percent of $\Delta<0$ and improve the average $\Delta$.} 
\label{tab:delta_0_percent}
\end{table}

\subsection{Analysis}
\label{sec:analysis}
\paragraph{\emph{Margin} between the NMT and the LM.}
Firstly, we analyze the distribution of the \emph{Margin} between the NMT and the LM (i.e., $\Delta$ in Eq.~\ref{eq:delta_def}). As shown in Figure~\ref{fig:delta_hist}, for the joint training model NMT+LM, although most of the \emph{Margin}s are positive, there are still many tokens with negative \emph{Margin} and a large amount of \emph{Margin}s around 0. 
This indicates that the LM is probably overconfident for many tokens, and addressing the overconfidence problem is meaningful for NMT. By comparison, the \emph{Margin} distribution of MSO is dramatically different with NMT+LM: the tokens with \emph{Margin} around 0 are significantly reduced, and the tokens with \emph{Margin} in $[0.75, 1.0]$ are increased apparently.

More precisely, we list the percentage of tokens with negative \emph{Margin} and the average \emph{Margin} for each model in Table \ref{tab:delta_0_percent}. Compared with NMT+LM, MTO and MSO reduce the percentage of negative \emph{Margin} by 2.28 and 1.56 points, respectively. We notice MSO performs slightly worse than MTO, because MSO neglects the hallucinations during training. 
As there are many tokens with negative \emph{Margin} in hallucinations, the ability of MSO to reduce the proportion of $\Delta<0$ is weakened. We further analyze effects of MTO and MSO on the average of \emph{Margin}. Both MTO and MSO improve the average of the \emph{Margin} by 33\% (from 0.33 to 0.44). In conclusion, MTO and MSO both indeed increase the \emph{Margin} between the NMT and the LM.

\begin{table}[tbp!]
\centering
\scalebox{1.0}{
\begin{tabular}{l | c | c}
\hline
\toprule
\textbf{Function} & \textbf{BLEU} & $\uparrow$ \\  \hline
NMT (Transformer)  & 25.75  & ref \\
~~~~+ LM & 25.90 & +0.15\\ \hline
~~~~~~~+ \emph{Linear}    &  26.13 & +0.38\\
~~~~~~~+ \emph{Cube}        &  26.45 & +0.60 \\
~~~~~~~+ \emph{Quintic}   & \textbf{26.94} & \textbf{+1.19} \\
~~~~~~~+ \emph{Log} ($\alpha=5$)  & 26.12 & +0.37 \\
~~~~~~~+ \emph{Log} ($\alpha=10$)  & 26.07 & +0.32 \\
~~~~~~~+ \emph{Log} ($\alpha=20$)  & 26.24 & +0.49 \\
\bottomrule
\end{tabular}}
\caption{Case-sensitive BLEU scores (\%) on Zh$\to$En test set of MTO with several variations of $\mathcal{M}(\Delta)$. $\alpha$ is the hyperparameter of \emph{Log}. All four $\mathcal{M}(\Delta)$ achieve BLEU improvements compared with NMT and NMT+LM, and \emph{Quintic} performs the best.}
\label{table:dif_margin_loss}
\end{table}

\paragraph{Variations of $\mathcal{M}(\Delta)$.}
We compare the performance of the four \emph{Margin} functions $\mathcal{M}(\Delta)$ defined in Section~\ref{sec:token_level}. We list the BLEU scores of the Transformer baseline, NMT+LM and our MTO approach with the four $\mathcal{M}(\Delta)$ in Table \ref{table:dif_margin_loss}. All the four variations bring improvements over NMT and NMT+LM. The results of \emph{Log} with different $\alpha$ are similar to \emph{Linear}, while far lower than \emph{Cube} and \emph{Quintic}. And \emph{Quintic} performs the best among all the four variations. We speculate the reason is that $\Delta \in [-0.5, 0.5]$ is the main range for improvement, and \emph{Quintic} updates more careful on this range (i.e., with smaller slopes) as shown in Figure~\ref{fig:nonlinear}.

\begin{table}[tbp!]
\centering
\scalebox{1.0}{
\begin{tabular}{l | c c}
\hline
\toprule
\textbf{Models} & \textbf{Valid}  & \textbf{Test} \\  \hline
NMT (Transformer)   & 23.67 & 25.75 \\
~~~~+ LM        & 23.61 & 25.90 \\
~~~~~~~+ MTO w/ Weight     & 24.09 & 26.94 \\
~~~~~~~+ MTO w/o Weight  & 23.36 & 25.85 \\
\bottomrule
\end{tabular}}
\caption{Case-sensitive BLEU scores (\%) on Zh$\to$En validation set and test set of MTO with (w/) and without (w/o) the weight $1 - p_{_{NMT}}(t)$.} 
\label{table:noweight_main}
\end{table}

\paragraph{Effects of the Weight of $\mathcal{M}(\Delta)$.}
In MTO, we propose the weight $1 - p_{_{NMT}}(t)$ of the \emph{Margin} function $\mathcal{M}(\Delta)$ in Eq.~\ref{eq:margin_term}. To validate the importance of it, we remove the weight and the \emph{Margin} loss degrades to $\mathcal{L}_{M} = \sum_{t=1}^{T} \mathcal{M}(\Delta(t))$. The results are listed in Table~\ref{table:noweight_main}. Compared with NMT+LM, MTO without weight performs worse with 0.25 and 0.05 BLEU decreases on the validation set and test set, respectively. Compared with MTO with weight, it decreases 0.73 and 1.09 BLEU scores on the validation set and test set, respectively. This demonstrates that the weight $1 - p_{_{NMT}}(t)$ is indispensable for our approach.

\begin{figure}[tbp!]
    \includegraphics[width=0.48\textwidth]{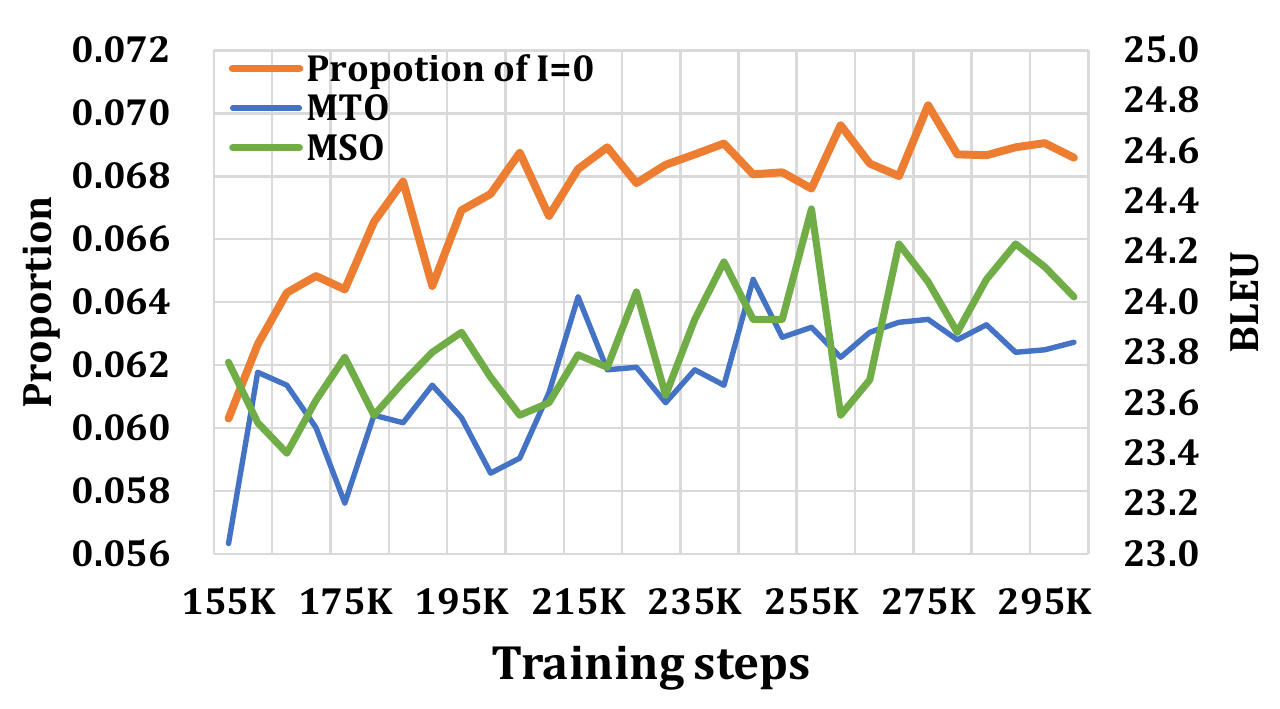}
    \caption{Changes of the proportion of $I_{\mathcal{R} (\mathbf{x}, \mathbf{y})< 30\%} = 0$ on Zh$\to$En during finetuning for MSO, and BLEU scores (\%) on the validation set of Zh$\to$En for MTO and MSO. The orange line corresponds to the left y-axis, and the green and blue lines correspond to the right y-axis. We sample 100K sentence pairs in the training data and compute $I_{\mathcal{R} (\mathbf{x}, \mathbf{y})< 30\%}$.} 
    \label{fig:weight_fn_trend}
\end{figure}

\paragraph{Changes of $I_{\mathcal{R} (\mathbf{x}, \mathbf{y})< k}$ During Training.}
In MSO, we propose a dynamic weight function $I_{\mathcal{R} (\mathbf{x}, \mathbf{y})< k}$ in Eq.~\ref{eq:sentence-loss}. Figure \ref{fig:weight_fn_trend} shows the changes of $I_{\mathcal{R} (\mathbf{x}, \mathbf{y})< k}$ in MSO and the BLEU scores of MSO and MTO during finetuning. As the training continues, our model gets more competent, and the proportion of sentences judged to be ``dirty data'' by our model increases rapidly at first and then flattens out, which is consistent with the trend of BLEU of MSO. Moreover, by adding the dynamic weight function, MSO outperforms MTO at most steps.

\paragraph{Case Study.} 
To better illustrate the translation quality of our approach, we show several translation examples in Appendix~\ref{sec:appendix_casestudy}. 
Our approach grasps more segments of the source sentences, which are mistranslated or neglected by the Transformer.

\section{Related Work}
\paragraph{Translation Adequacy of NMT.}
NMT suffers from the hallucination and inadequacy problem for a long time~\cite{tu2016modeling,muller-etal-2020-domain, hallucination_2020_acl, lee2019hallucinations}.
Many studies improve the architecture of NMT to alleviate the inadequacy issue, including tracking translation adequacy by coverage vectors~\cite{tu2016modeling, mi-etal-2016-coverage}, modeling a global representation of source side~\cite{weng2020gret}, dividing the source sentence into past and future parts~\cite{zheng-etal-2019-dynamic}, and multi-task learning to improve encoder and cross-attention modules in decoder~\cite{meng2016interactive,meng2018neural,weng-etal-2020-towards}. They inductively increase the translation adequacy, while our approaches directly maximize the \emph{Margin} between the NMT and the LM to prevent the LM from being overconfident. 
Other studies enhance the translation adequacy by adequacy metrics or additional optimization objectives. \citet{Tu:reconstruct:2017:AAAI} minimize the difference between the original source sentence and the reconstruction source sentence of NMT. \citet{Kong_Tu_Shi_Hovy_Zhang_2019} propose a coverage ratio of the source sentence by the model translation. \citet{feng2019fluency_faithfulness} evaluate the fluency and adequacy of translations with an evaluation module. However, the metrics or objectives in the above approaches may not wholly represent adequacy. On the contrary, our approaches are derived from the criteria of the NMT model and the LM, thus credible.

\paragraph{Language Model Augmented NMT.}
Language Models are always used to provide more information to improve NMT. For low-resource tasks, the LM trained on extra monolingual data can re-rank the translations by fusion~\cite{shallow-deep-fusion:2015, cold_fusion, simple_fusion_2018}, enhance NMT's representations~\cite{ clinchant-etal-2019-bert, Zhu2020Incorporating-bert}, and provide prior knowledge for NMT~\cite{lm_prior:2020:EMNLP}. For data augmentation, LMs are used to replace words in sentences~\cite{kobayashi-2018-contextual, wu2018conditional-context, gao-2019-soft-context}. Differently, we mainly focus on the \emph{Margin} between the NMT and the LM, and no additional data is required. \citet{simple_fusion_2018} propose the Simple Fusion approach to model the difference between NMT and LM. Differently, it is trained to optimize the residual probability, positively correlated to $p_{_{NMT}}/p_{_{LM}}$ which is hard to optimize and the LM is still required in inference, slowing down the inference speed largely. 

\paragraph{Data Selection in NMT.}
Data selection and data filter methods have been widely used in NMT. To balance data domains or enhance the data quality generated by back-translation~\cite{back-translation-sennrich-etal-2016-neural}, many approaches have been proposed, such as utilizing language models~\cite{moore-lewis-2010-intelligent, van-der-wees-etal-2017-dynamic, zhang-etal-2020-parallel-corpus}, translation models~\cite{junczys-dowmunt-2018-dual, wang-etal-2019-improving}, and curriculum learning~\cite{zhang-etal-2019-curriculum, wang2019dynamically}. Different from the above methods, our MSO dynamically combines language models with translation models for data selection during training, making full use of the models.

\section{Conclusion}
We alleviate the problem of inadequacy translation from the perspective of preventing the LM from being overconfident. Specifically, we firstly propose an indicator of the overconfidence degree of the LM in NMT, i.e., \emph{Margin} between the NMT and the LM. Then we propose \emph{Margin}-based Token-level and Sentence-level objectives to maximize the \emph{Margin}. Experimental results on three large-scale translation tasks demonstrate the effectiveness and superiority of our approaches.
The human evaluation further verifies that our methods can improve translation adequacy and fluency. 

\section*{Acknowledgments}
The research work descried in this paper has been supported by the National Nature Science Foundation of China (No. 12026606). The authors would like to thank the anonymous reviewers for their valuable comments and suggestions to improve this paper.

\bibliography{acl2021}
\bibliographystyle{acl_natbib}

\appendix

\section{Loss of the Language Model}
\label{sec:lm_loss}
\begin{figure}[th!]
    \centering
    \includegraphics[width=0.5\textwidth]{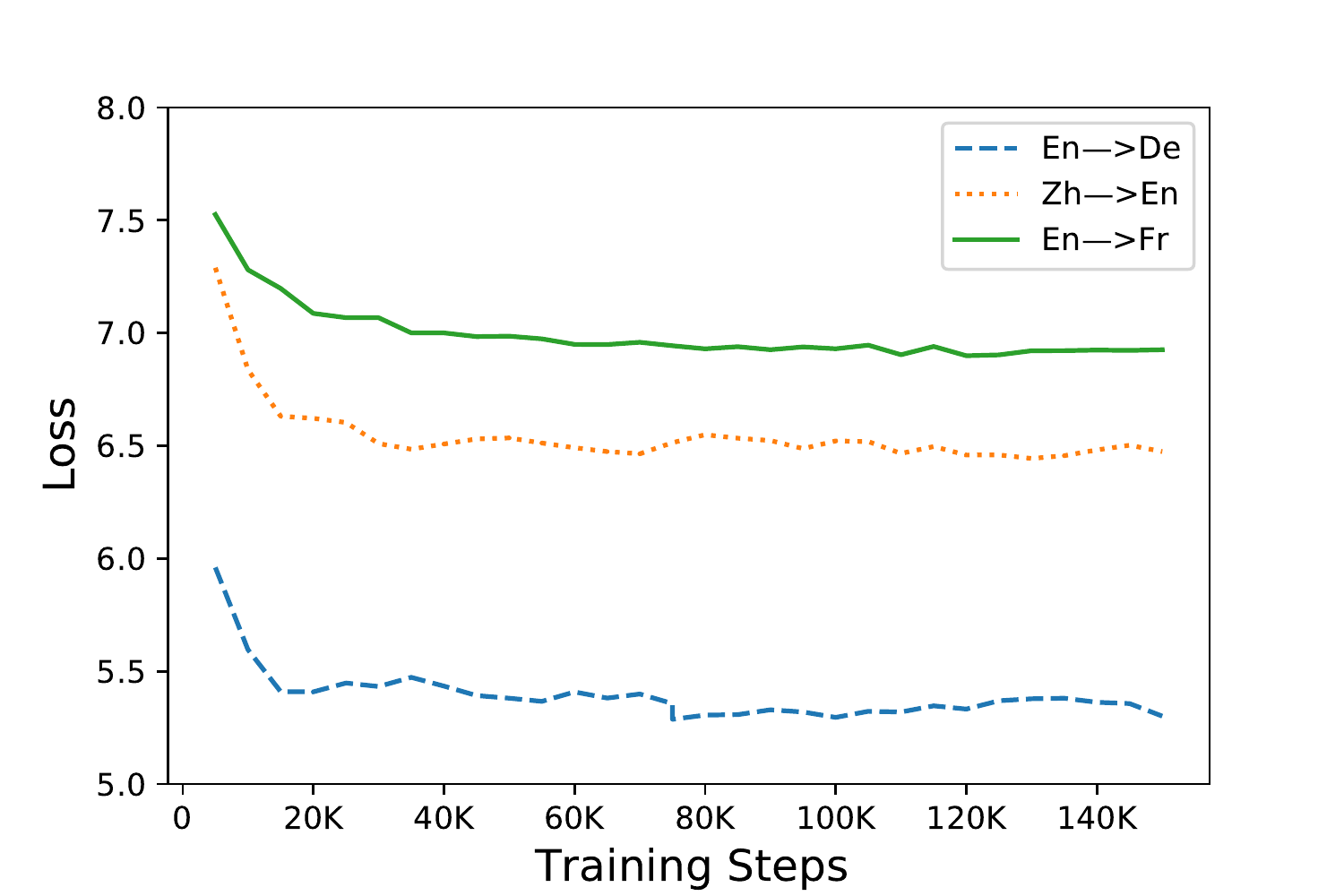}
    \caption{The loss of the LM on the validation set during pretraining for En$\to$De, Zh$\to$En and En$\to$Fr. The LM converges after training nearly 80K steps for all the three tasks.}
    \label{fig:lm_loss}
\end{figure}

To validate whether the LM is converged or not after pretraining, we plot the loss of the LM as shown in Figure~\ref{fig:lm_loss}. The loss of the LM remains stable after training about 80K steps for En$\to$De, Zh$\to$En and En$\to$Fr, indicating that the LM is converged during pretraining stage.

\begin{figure}[t!]
    \centering     
    \subfigure[$\lambda_{M}$ (En$\to$De)]{
        \label{fig:ende_lambda_margin}
        \begin{minipage}{.45\columnwidth}
        \includegraphics[width=\columnwidth]{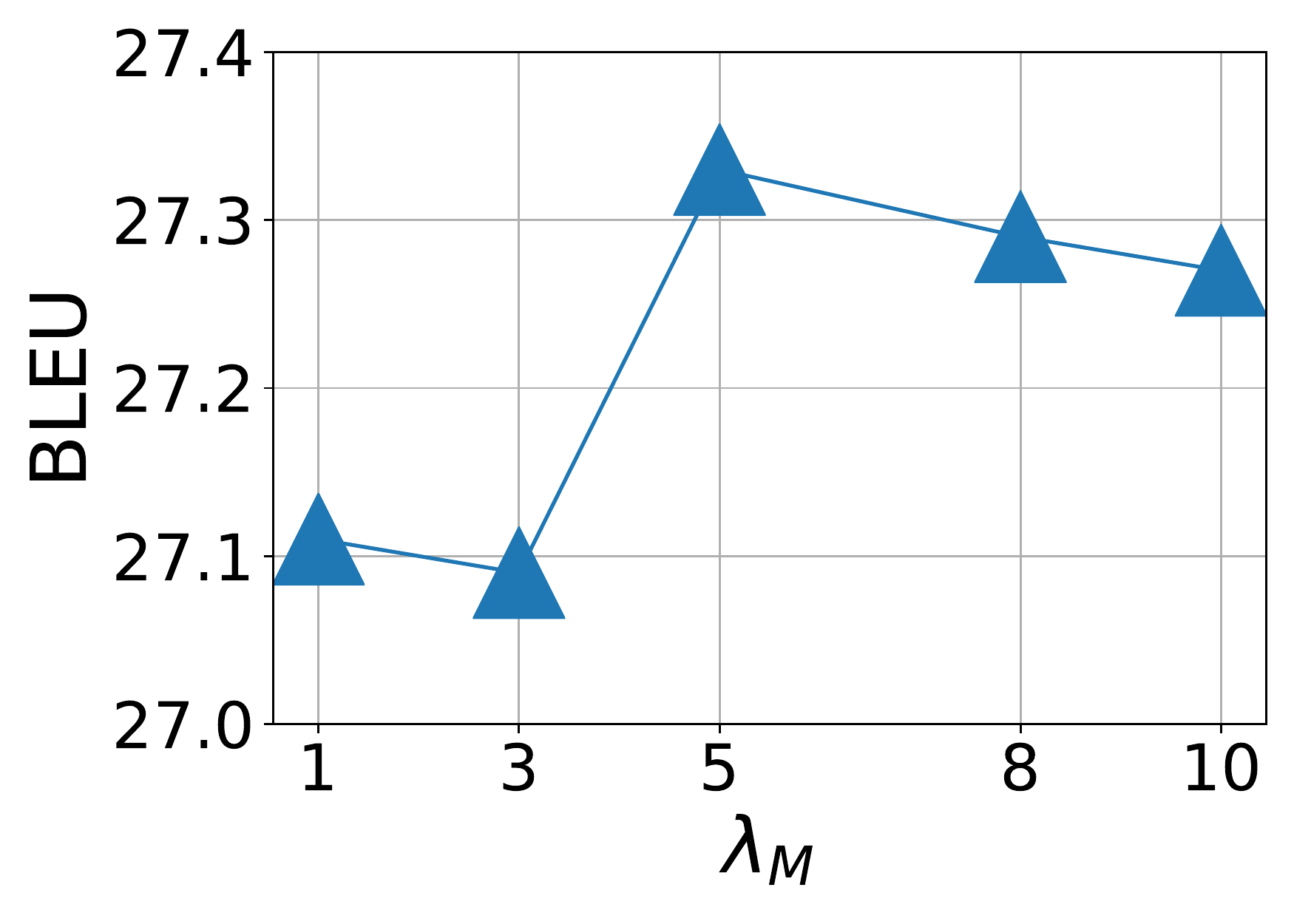}
        \end{minipage}
    }
    \subfigure[$k$ (En$\to$De)]{
        \label{fig:ende_sent_threshold}
        \begin{minipage}{.45\columnwidth}
        \includegraphics[width=\columnwidth]{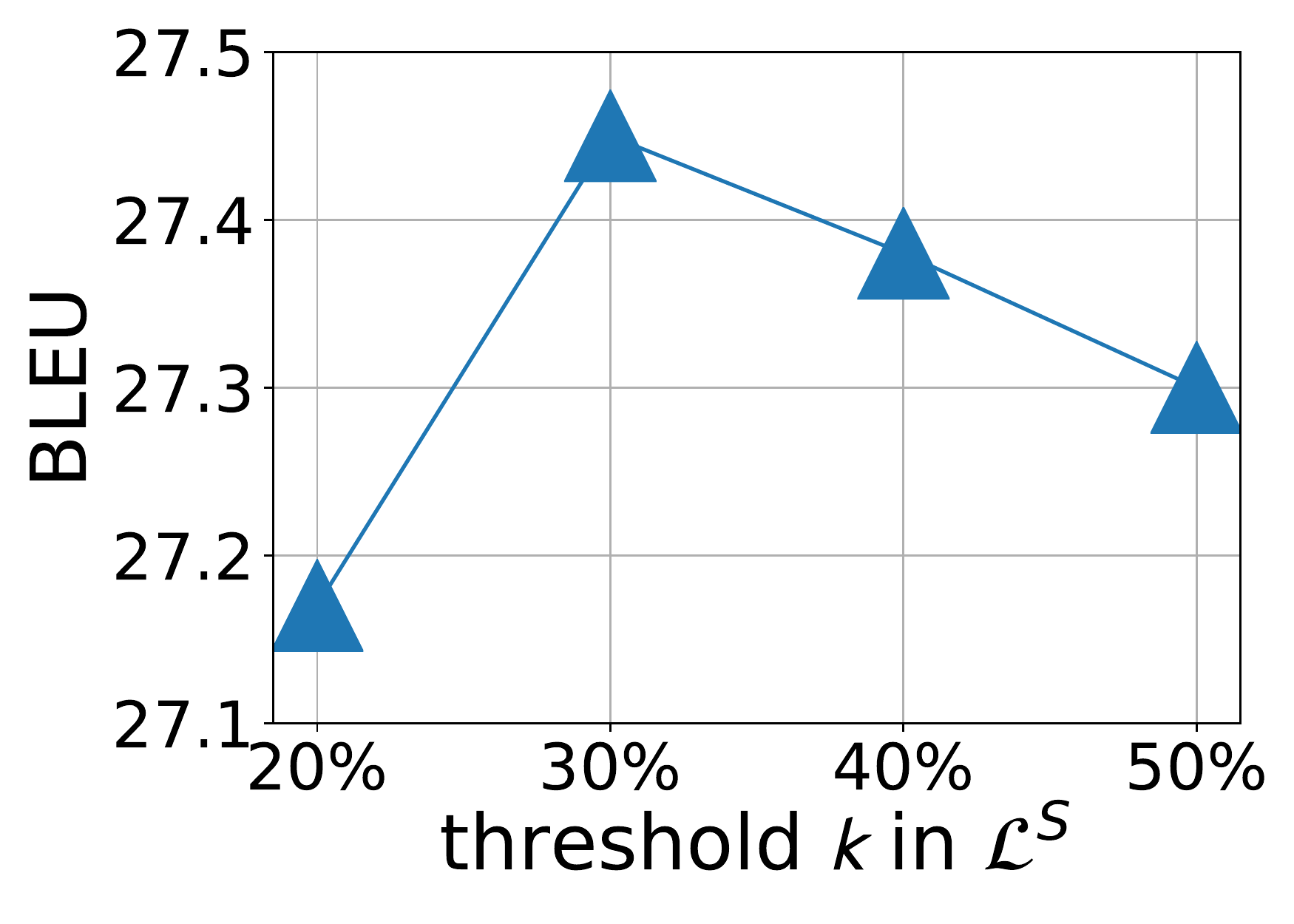}
        \end{minipage}
        }

    \subfigure[$\lambda_{M}$ (En$\to$Fr)]{
        \label{fig:enfr_lambda_margin}
        \begin{minipage}{.45\columnwidth}
        \includegraphics[width=\columnwidth]{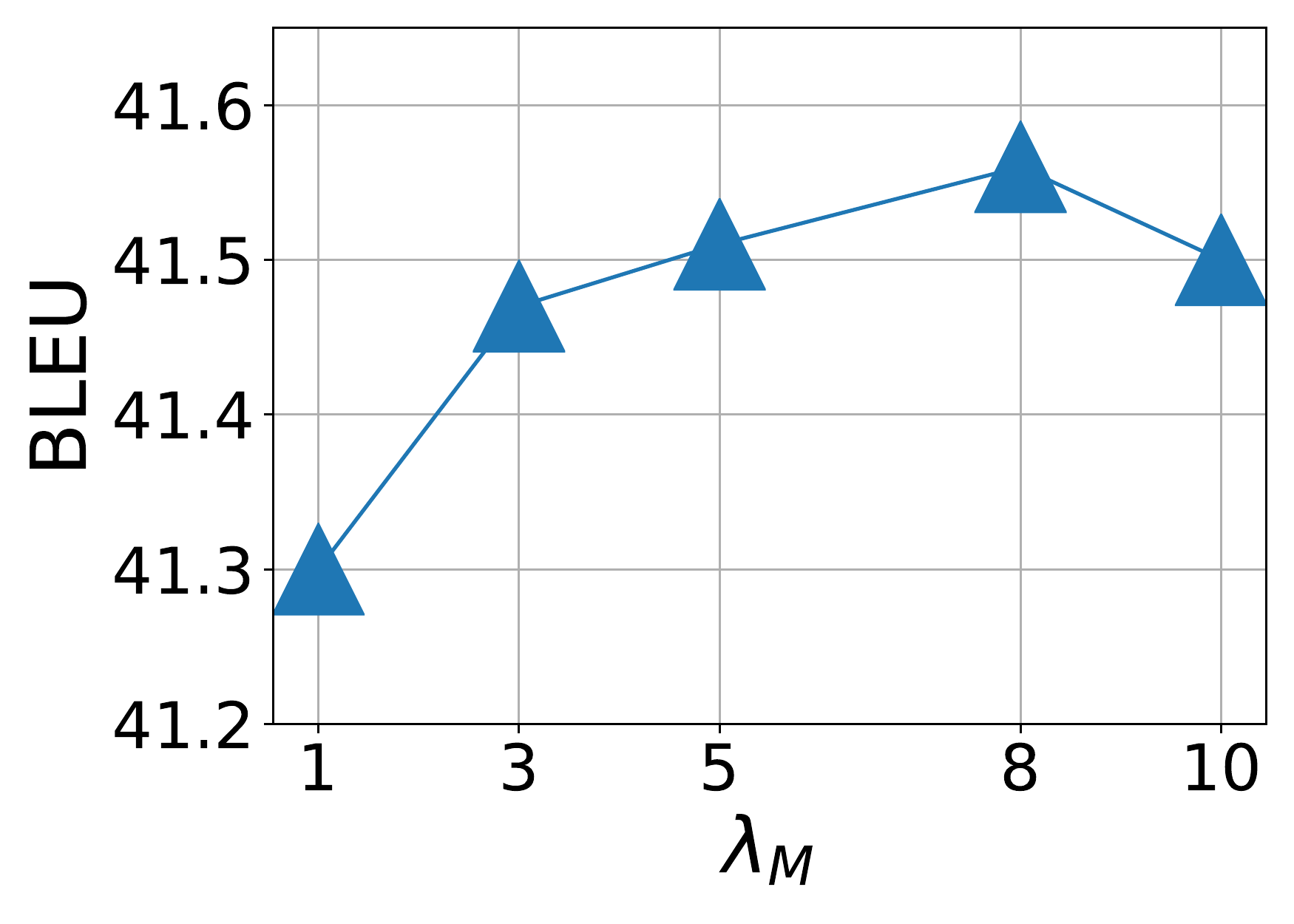}
        \end{minipage}
    }
    \subfigure[$k$ (En$\to$Fr)]{
        \label{fig:enfr_sent_threshold}
        \begin{minipage}{.45\columnwidth}
        \includegraphics[width=\columnwidth]{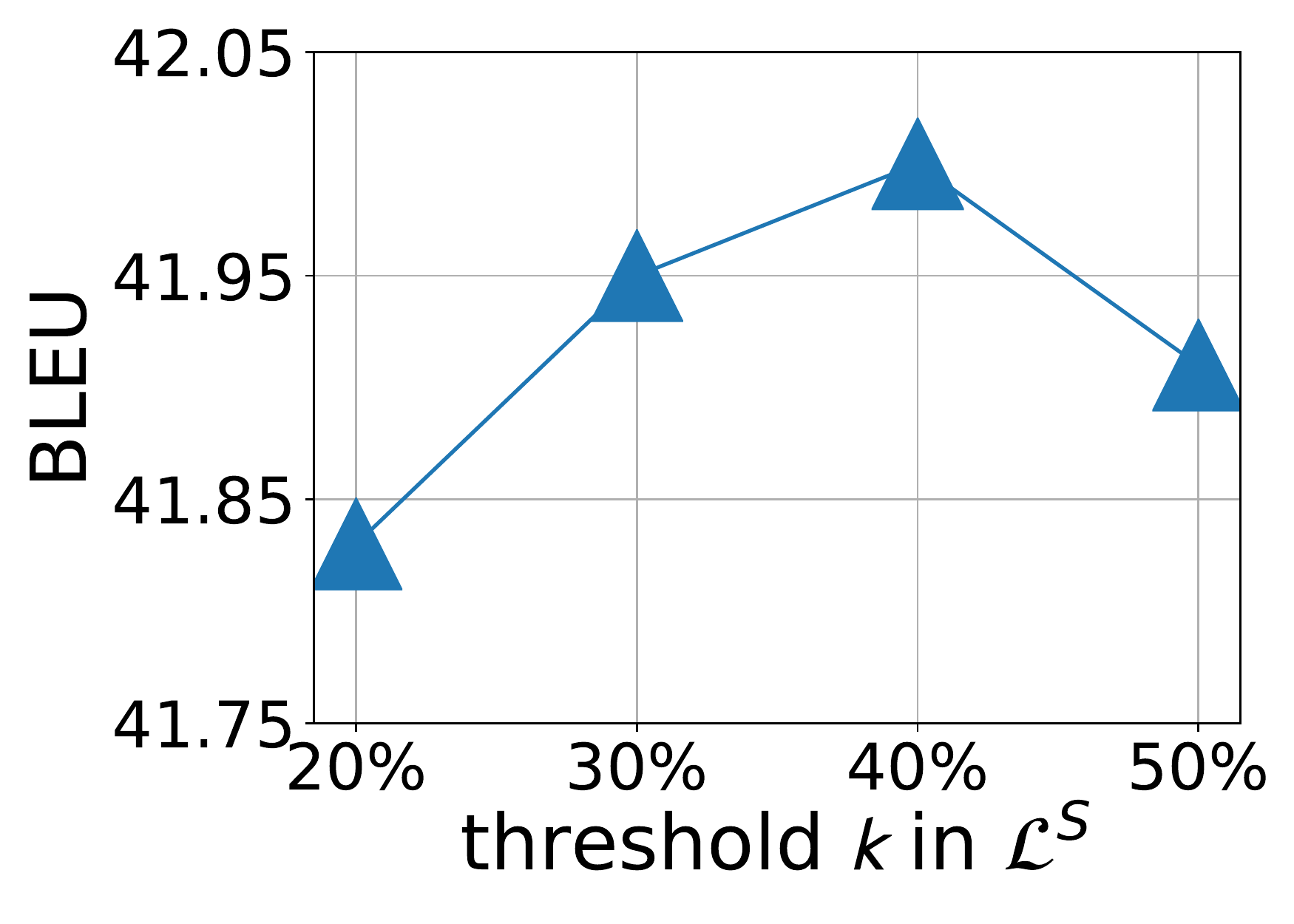}
        \end{minipage}
        }

    \subfigure[$\lambda_{M}$ (Zh$\to$En)]{
        \label{fig:zhen_lambda_margin}
        \begin{minipage}{.45\columnwidth}
        \includegraphics[width=\columnwidth]{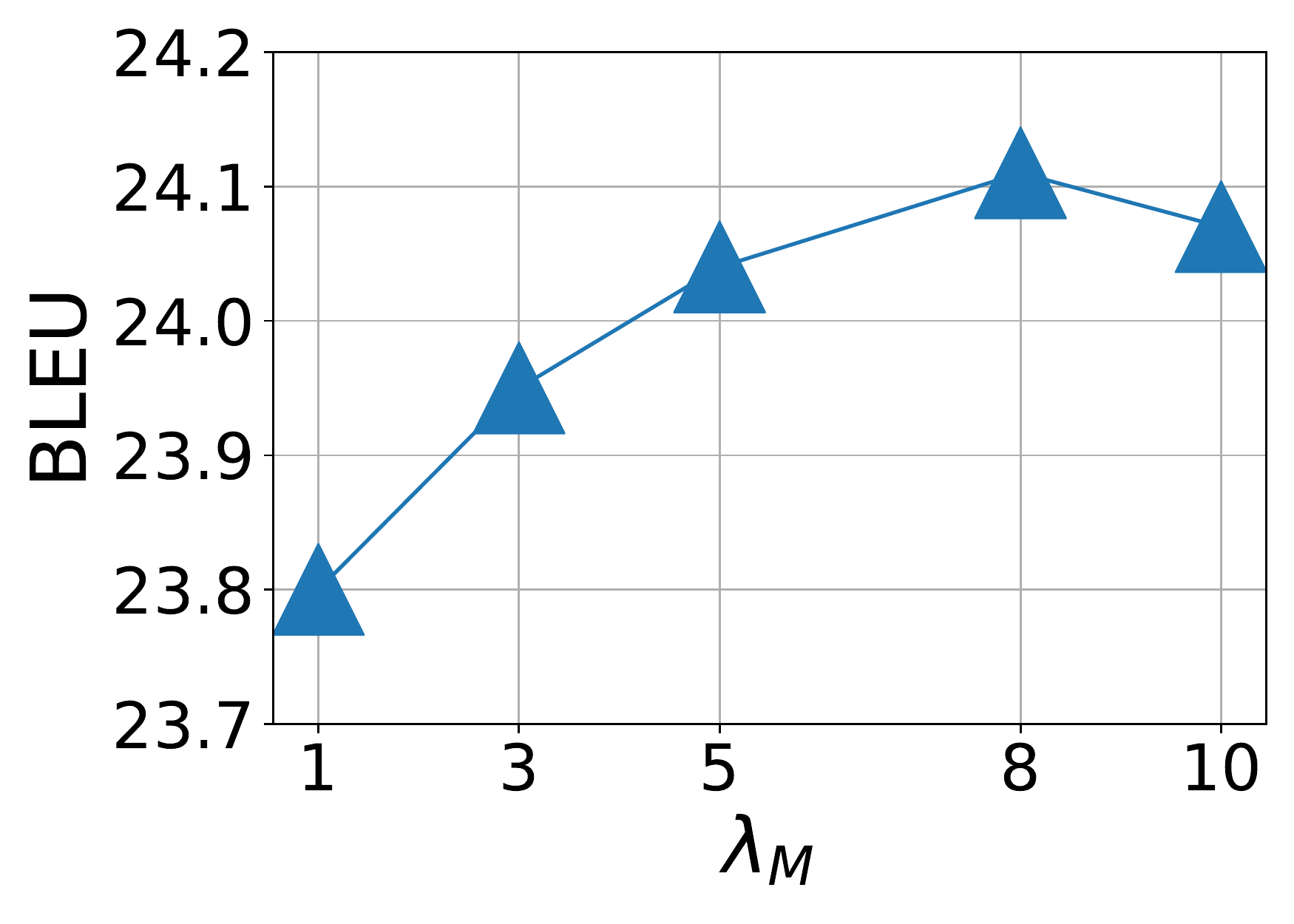}
        \end{minipage}
    }
    \subfigure[$k$ (Zh$\to$En)]{
        \label{fig:zhen_sent_threshold}
        \begin{minipage}{.45\columnwidth}
        \includegraphics[width=\columnwidth]{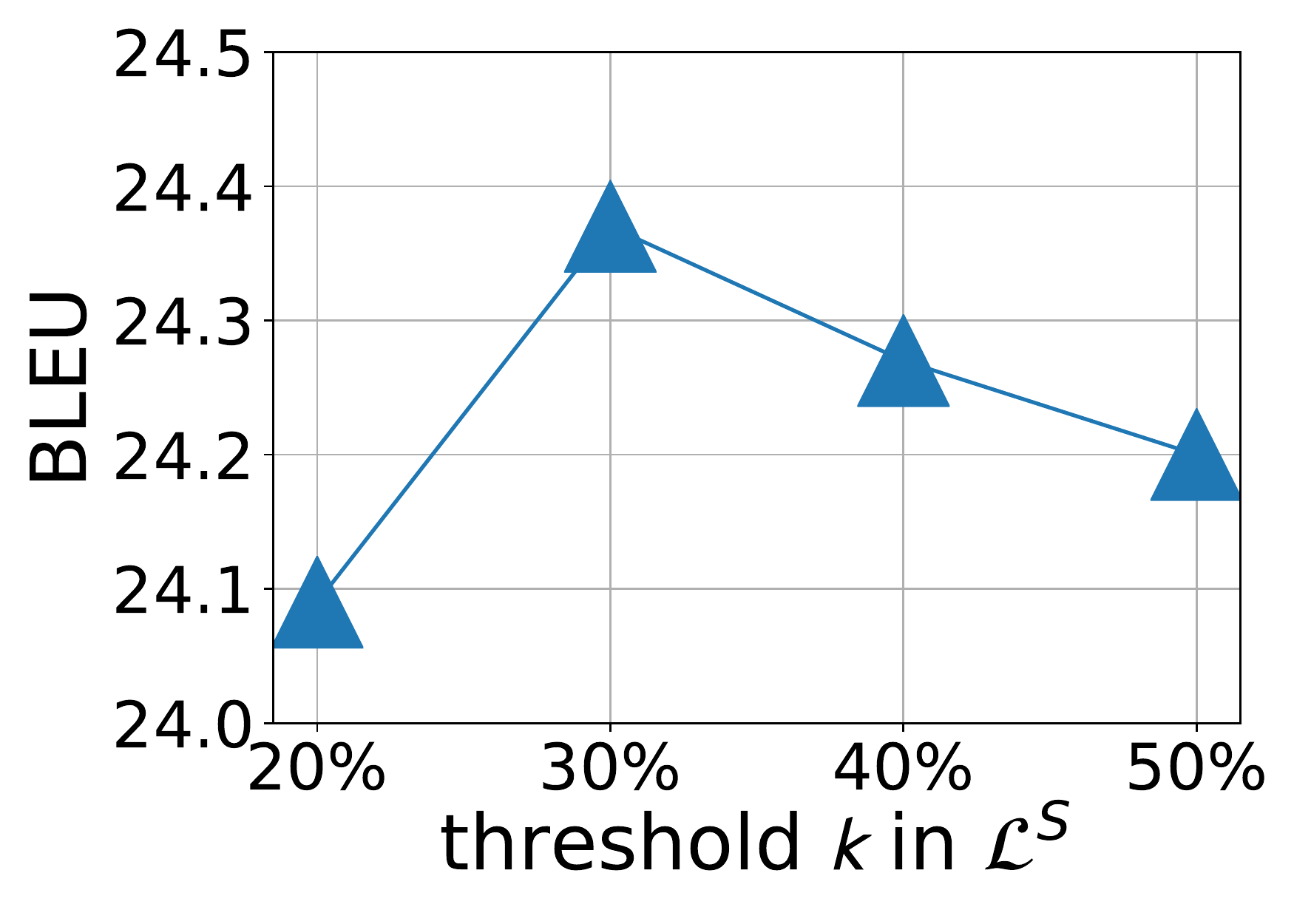}
        \end{minipage}
        }
    \caption{Case-sensitive BLEU scores (\%) on validation sets of WMT14 En$\to$De, WMT14 En$\to$Fr and WMT19 Zh$\to$En with different hyperparameters, respectively. $\lambda_{M}$ is defined in Eq.~\ref{eq:token_loss}, and the search results are shown in Figure (a), (c) and (e). The threshold $k$ for MSO is defined in Eq.~\ref{eq:sentence-loss} and the results of it are shown in Figure (b), (d), and (f).}    
    \label{fig:hyper_selection}
\end{figure}

\section{Hyperparameters Selection}
\label{sec:hyper-select}
The results of our approaches with different $\lambda_{M}$ (defined in Eq.~\ref{eq:token_loss}) and $k$ (defined in Eq.~\ref{eq:sentence-loss}) on the validation sets of WMT14 En$\to$De, WMT14 En$\to$Fr and WMT19 Zh$\to$En are shown in Figure~\ref{fig:hyper_selection}. We firstly search the best $\lambda_M$ based on MTO. All the three datasets achieve better performance for $\lambda_M\in [5,10]$. The model reaches the peak when $\lambda_{M}=$5, 8, and 8 for the three tasks, respectively. Then, fixing the best $\lambda_M$ for each dataset, we search the best threshold $k$. As shown in the right of Figure~\ref{fig:hyper_selection}, the best $k$ is 30\% for En$\to$De and Zh$\to$En, 40\% for En$\to$Fr. This is consistent with our observations. When the proportion of tokens with negative \emph{Margin} in a target sentence is greater than 30\% or 40\%, the sentence is most likely to be a hallucination.

\begin{figure*}[!ht]
    \centering
    \includegraphics[width=0.95\textwidth]{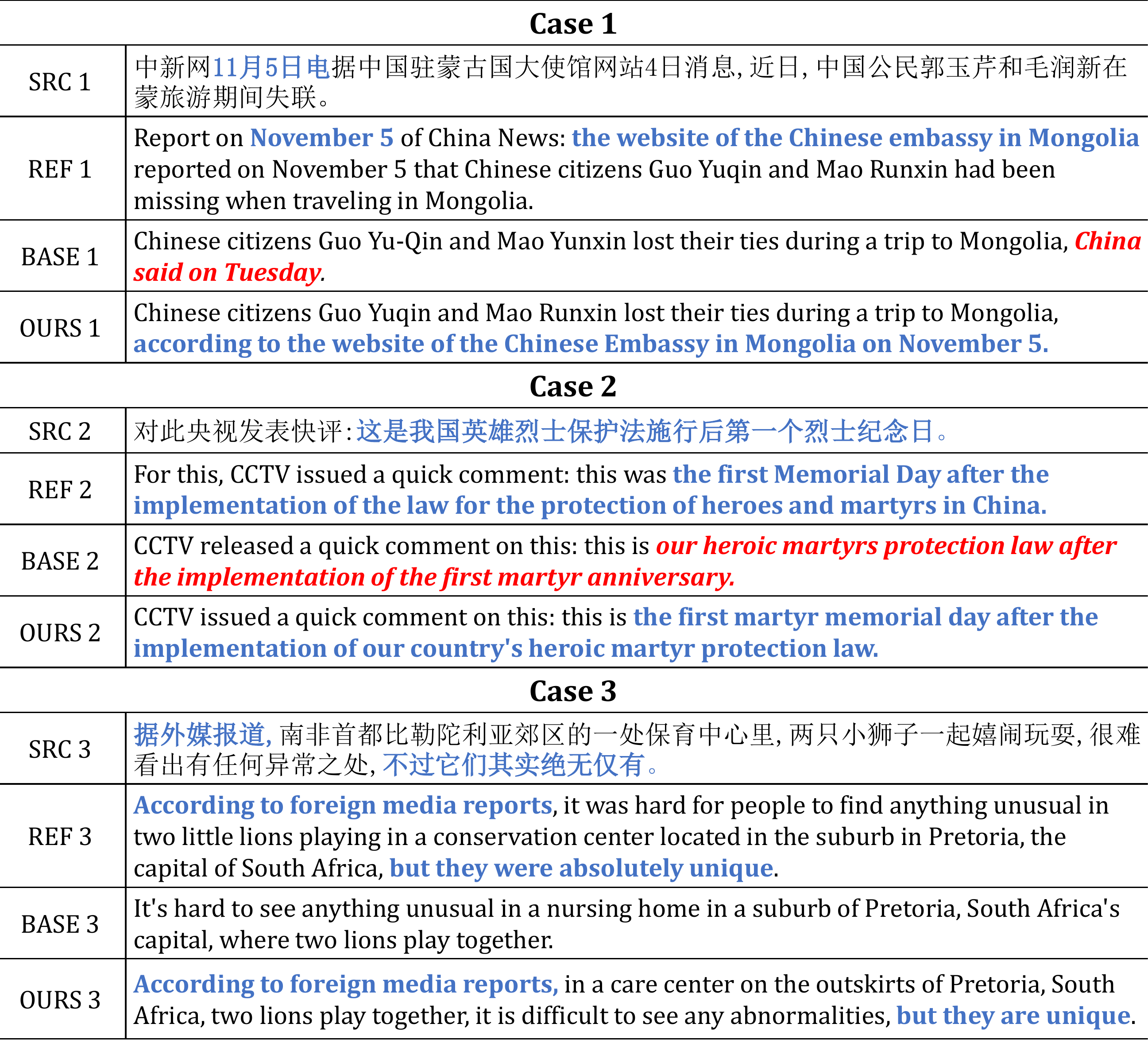}
    \caption{Several example sentence pairs (SRC, REF) from WMT19 Zh$\to$En test set. We list the translation of the Transformer baseline (BASE) and our MSO method (OURS). The text in bold red font is mistranslated by the base model. The text in bold blue font is mistranslated or under-translated by the base model but translated correctly by our model.}
    \label{fig:case_study}
\end{figure*}

\section{Case Study} \label{sec:appendix_casestudy}
As shown in Figure~\ref{fig:case_study}, our approach outperforms the base model (i.e., the Transformer) in translation adequacy. In case 1, the base model generates ``on Tuesday'', which is unrelated to the source sentence, i.e., hallucination, and under-translates ``November 5'' and ``the website of the Chinese embassy in Mongolia'' information in the source sentence. However, our approach translates the above two segments well. In Case 2, the base model reverses the chronological order of the source sentence, thus generates a mis-translation, while our model translates perfectly. In Case 3, the base model neglects two main segments of the source sentence (the text in bold blue font) and leads to the inadequacy problem. However, our model takes them into account. According to the three examples, we conclude that our approach alleviates the inadequacy problem which is extremely harmful to NMT.

\end{document}